\documentclass{article}

\usepackage[preprint]{neurips_2023}

\PassOptionsToPackage{numbers}{natbib}
\usepackage[utf8]{inputenc} 
\usepackage[T1]{fontenc}    
\usepackage{hyperref}       
\usepackage{url}            
\usepackage{booktabs}       
\usepackage{amsfonts}       
\usepackage{nicefrac}       
\usepackage{microtype}      
\usepackage{xcolor}         
\usepackage{graphicx}
\usepackage{adjustbox}

\title{Benchmarking pre-trained text embedding models in aligning built asset information}

\author{%
  Mehrzad Shahinmoghadam\thanks{For correspondence, please contact: \texttt{mehrzad.shahinmoghadam.1@ens.etsmtl.ca}} \\
  Department of construction engineering  \\
  École de technologie supérieure \\
  Montreal, H3C 1K3, Canada \\
  \texttt{mehrzad.shahinmoghadam.1@ens.etsmtl.ca} \\
  \And
  Ali Motamedi \\
  Department of construction engineering  \\
  École de technologie supérieure \\
  Montreal, H3C 1K3, Canada \\
  \texttt{ali.motamedi@etsmtl.ca} \\
}

\begin{document}

\maketitle

\begin{abstract}
  Accurate mapping of the built asset information to established data classification systems and taxonomies is crucial for effective asset management, whether for compliance at project handover or ad-hoc data integration scenarios. Due to the complex nature of built asset data, which predominantly comprises technical text elements, this process remains largely manual and reliant on domain expert input. Recent breakthroughs in contextual text representation learning (text embedding), particularly through pre-trained large language models, offer promising approaches that can facilitate the automation of cross-mapping of the built asset data. However, no comprehensive evaluation has yet been conducted to assess these models' ability to effectively represent the complex semantics specific to built asset technical terminology. This study presents a comparative benchmark of state-of-the-art text embedding models to evaluate their effectiveness in aligning built asset information with domain-specific technical concepts. Our proposed datasets are derived from two renowned built asset data classification dictionaries. The results of our benchmarking across six proposed datasets, covering three tasks of clustering, retrieval, and reranking, highlight the need for future research on domain adaptation techniques. The benchmarking resources are published as an open-source library, which will be maintained and extended to support future evaluations in this field. 
\end{abstract}

\section{Introduction}
Asset management plays a pivotal role in ensuring optimal performance and extended life span of the built environment through a systematic process of monitoring and maintaining various facilities and equipment. The rapid advancement of digital technologies has led asset owners to increasingly demand enriched digital twins at project handover to support real-time operations and maintenance of the built assets \citep{love2019benefits}. Simultaneously, the growing awareness of the benefits of digitized asset management highlights the essential need for federated access to built asset data \citep{moretti2023federated}. This requires aligning extensive data sources and their underlying schema with established data models, classification systems, or taxonomies to facilitate data accessibility for diverse stakeholders and improve interoperability across various software environments. However, aligning built asset data with pre-defined classification systems poses significant challenges in practice. A key challenge stems from the multi-source and multi-disciplinary nature of built asset data, which leads to the use of diverse formats and terminologies across different projects and stakeholders. For example, the terminology that architects utilize to describe the specifications for a particular building component or system can vastly differ from those used by structural engineers or subcontractors. Moreover, the structures of domain-specific classifications used in different disciplines often vary in granularity. For instance, the detailed engineering descriptions of an HVAC system provided by mechanical engineers may be far more comprehensive than those required and used by operations and maintenance teams. Finally, variations in local regulations and standards can further complicate the alignment process, particularly for large-scale or international projects. These issues, combined with the dynamic and evolving nature of built asset data throughout an asset's lifecycle, lead to potential inconsistencies when integrating this data into a unified digital asset management environment. 

In response, there have been several initiatives aimed at facilitating the digital delivery of built asset information while ensuring its conformity with predefined or standardized descriptions (data models, taxonomies, etc.). One major initiative is buildingSMART Data Dictionary (bSDD)\citep{bsDD2024}, an international and ongoing effort whose main objective is to create shared definitions for describing the built environment. This is achieved through a collection of interconnected data dictionaries that are both human-readable and machine-readable\citep{bsDD2024}. Although making various data dictionaries programmatically accessible will facilitate access to agreed and consistent terms, the complexity and dynamic diversity of the built asset terminology necessitate robust data mapping strategies to accommodate various data descriptions and updates \citep{forth2024domain}. As a result, the asset information alignment process remains predominantly manual, relying heavily on the expertise of domain specialists to accurately map complex technical data \citep{roberts2018digitalising}. The significant challenges associated with the manual alignment process, including high costs, time consumption, and potential for human error, highlight the need for more automated and reliable data mapping solutions.

The central thesis of our research builds upon the argument that recent advancements in natural language processing/understanding research can significantly enhance automated data mapping processes. In particular, the rich and contextualized representation of textual inputs as numeric vectors, commonly known as text embedding \citep{pennington2014glove,lee2024gecko}, provides advanced capabilities for machines to understand the semantics of the intricate terminologies. Earlier methods such as word2vec \citep{mikolov2013distributed} and GloVe \citep{pennington2014glove} relied on static embeddings, i.e., generating fixed representations of numerical vectors for each word based on their co-occurrence in large corpora. However, recent neural language models, dominantly built on top of the transformer architecture \citep{vaswani2017attention}, can generate dynamic, context-sensitive embeddings. The capability of recent embedding models in adapting the representation of words (or sub-word tokens) based on their surrounding context has motivated researchers and practitioners across diverse fields to leverage the power of contextual text embeddings to drive advancements in their respective domains. From traditional databases integration \citep{cappuzzo2020creating}, to information management in biomedicine \citep{zhang2019biowordvec}, or public figure perceptions in social science studies \citep{cao2024large}, the increasing volume of encouraging reports on leveraging text embedding models to deliver a more nuanced text understanding in various specialized domains \citep{rasmy2021med, ostendorff2021evaluating, rouhizadeh2024dataset, wilkho2024ff, cao2024large} reinforces the relevance of these models in automating data alignment in the domain of built asset information management.

Based on the observation that built asset data predominantly exists in textual form \citep{wu2022natural}, we argue that state-of-the-art text embedding models present promising opportunities to refine the automated alignment of built asset information. However, the extensive and increasing availability of pre-trained language models has led to the proliferation of potential text embedding models, creating confusion regarding model selection for different use cases \citep{muennighoff2022mteb}. Moreover, recent research indicates that general-purpose text embedding models often struggle to maintain consistent performance across diverse tasks and domains \citep{lee2024gecko}. This is while most previous studies utilizing pre-trained or fine-tuned language models in built environment research have been significantly limited in scope, primarily focusing on ad-hoc downstream tasks with small evaluation datasets \citep{shahinmoghadam2024neural, jung2024transformer, wang2024pre, forth2024domain, jeon2024dynamic}. Such limitations can result in a potentially skewed perspective on the overall domain-specific text understanding of these models \citep{shahinmoghadam2024neural}. Additionally, scarce public access to the datasets used in previous works poses another important challenge to the transparency and reproducibility of the reported results. This motivates us to examine the extent to which existing language models can be directly leveraged to deliver contextually accurate mappings of domain-specific terminology within the context of built asset information management. In this work, we present a comprehensive benchmark of state-of-the-art text embedding models to evaluate their effectiveness in capturing and representing the semantics of textual descriptions related to built assets. Through this evaluation, we aim to identify the strengths and limitations of existing language models in enhancing data alignment practices within the built asset domain. Our proposed benchmark is aligned with the Massive Text Embedding Benchmark (MTEB) \citep{muennighoff2022mteb}, a benchmark recognized extensively in both academic and practical contexts for its robustness and utility. We benchmark 24 text embedding models on our developed datasets that amount to a total of more than ten thousand data entries across six tasks, making our evaluations the most comprehensive ones in this specialized field to date. By making our datasets and benchmark software publicly available, we encourage future research to build upon our work, contributing to continuous improvements in this domain.

\section{Methods}
\subsection{Data sources}
Given the built environment's multidisciplinary nature, the datasets included in the benchmark must encompass an expansive spectrum of sub-domain subjects, including architectural, structural, mechanical, and electrical systems. To ensure a diverse coverage of built products in our benchmark, we carefully examined the selection of data sources used for creating task-specific datasets. A detailed description of the corpus development and data extraction processes is provided below.

The initial step in creating the benchmark's task-specific datasets is the development of a consistent corpus of built products. Based on the requirements of the tasks within our benchmark, the core corpus needed to include the following key information for each product: name or title, description, and corresponding labels (group categories). The two primary sources used to develop the built product corpora are as follows:

\textbf{Industry Foundation Classes (IFC).} Published and maintained by buildingSMART International\citep{bsi2024}, IFC is an open international data model offering comprehensive digital descriptions of various aspects of building and infrastructure projects. Originally designed to facilitate interoperability and information exchange among different software applications and stakeholders, IFC provides a comprehensive representation of various aspects of built asset entities. We utilize IFC version 4.3.2.0 \citep{ifc-doc2024}, recently approved as an ISO standard (ISO 16739-1:2024).

\textbf{Uniclass.} Developed and maintained by the National Building Specification (NBS)\citep{nbs2024}, Uniclass is a unified classification system for the built environment. We utilize version 1.33 of the Uniclass Pr Product Table\citep{uniclass2024}. Uniclass has extensive coverage, encompassing over 8,000 product types, making it one of the most recognized and widely adopted classification systems in the built asset industry.

\begin{figure}[ht]
\centering
\includegraphics[width=0.95\textwidth]{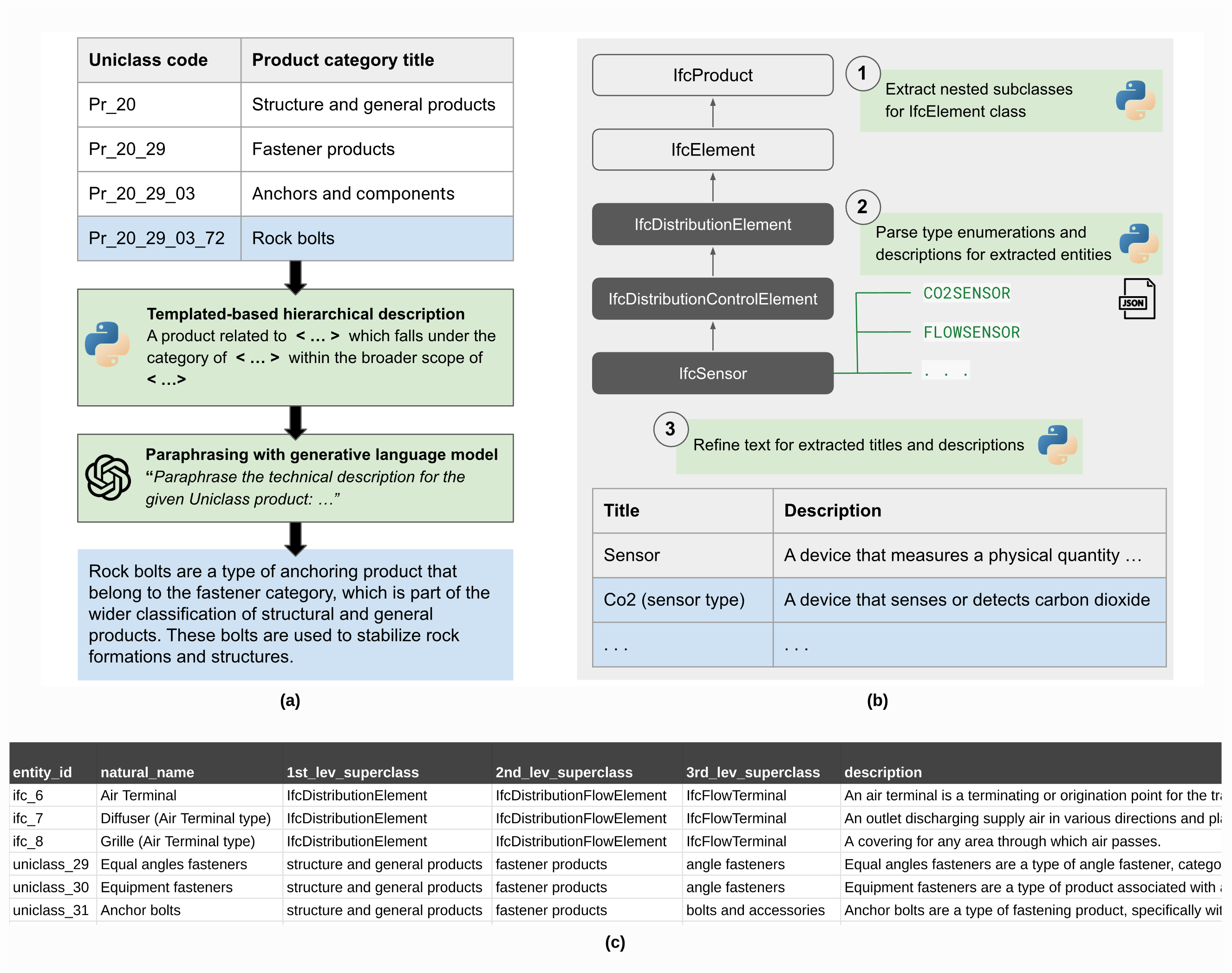}
\caption{Overview of the main steps in developing the built product corpus: (a) Example of extracting categories and synthesizing entity descriptions from raw Uniclass entries; (b) Example of hierarchical relation extraction for main entities and their enumerated types from the IFC schema; (c) Sample records from the developed corpus, containing product titles, descriptions, and categories with three levels of granularity.}
\label{fig:methods}
\end{figure}

\subsection{Data extraction}
To create a corpus of products with corresponding names, descriptions, and labels, we undertook the following steps: For Uniclass, we utilize the publicly-available CSV format of the products table, which comprises over 8,000 products categorized into three hierarchical levels. Product names were directly extracted from the table, while product categories were inferred from the numeric codes associated with hierarchical categories (see Figure \ref{fig:methods}). To automatically extract the corresponding textual labels for each product, we developed a script to scrape the table programmatically. As the original table does not include product descriptions, we propose a method (detailed in the subsequent subsection) to synthesize a description for each product. We retained only those products that have labels for all three classification levels. After applying this filtering process, the Uniclass corpus comprises 4,234 instances, which remains sufficiently large for our benchmarking purposes.

Regarding the IFC schema, we parse the official schema content by utilizing resources from an open-source Python library\citep{ifcopenshell2024} that enables programmatic access to IFC entities. Initially, we extracted entities of interest from a JSON-formatted file containing the complete list of IFC entities, their type enumeration, and their definition (derived from IFC's official documentation). An analysis of the "IfcProduct" class within the IFC schema indicated that a significant majority of product entities are classified under the "IfcElement" class. Therefore, we focused exclusively on the "IfcElement" subclasses. After removing IFC entities with missing descriptions (less than 1\% of total "IfcElement" entities), we developed a script to extract each entity's top three parent classes to serve as the product category labels. In addition to entity superclass groups, we use the domain-specific schemas (e.g., structural, HVAC, building control) from IFC's official documentations\citep{ifc-doc2024} as an additional source for entity label assignment. The resulting IFC corpus comprises 977 entities (total of parent entities and type enumerations).

\subsection{Data augmentation and curation}
The process of generating textual descriptions for Uniclass entities is depicted in Figure \ref{fig:methods}(a). Initial entity descriptions are synthesized by sequentially concatenating the entity's category titles, progressing from the most specific to the most general. An example of the synthesized descriptions is provided in Figure \ref{fig:methods}(a). These concatenated descriptions are then paraphrased using a generative language model to create more nuanced and natural descriptions, relaxing the text from the rigid template initially employed. We generated paraphrased descriptions using the most advanced version of the GPT-4 model available at the time of conducting the experiments (gpt-4-turbo-2024-04-09). Although the prompts used for generating paraphrased descriptions were designed to prevent the alteration or addition of facts, it was essential to manually review all generated descriptions due to known potential inaccuracies of generative language models. The review is carried out by two domain experts, each with over ten years of experience in the field. Each expert cross-checked the issues identified by the other, and the final decisions were made based on mutual agreement. 
The following curation steps are undertaken to ensure the accuracy and consistency of the extracted product names and descriptions. We preprocess native IFC entity names and convert them into a readable form (e.g., "IfcHeatExchanger" to Heat Exchanger; see examples in Figure \ref{fig:methods}(b) and (c)). For IFC class enumeration types, where the enumeration name alone might be ambiguous, we append the parent class type in parentheses. For example, the enumeration WATER, a subclass of "IfcBoilerTypeEnum", is represented as "WATER (Boiler Type)" (see examples in Figure \ref{fig:methods}(c)). Following the same logic, we enrich the product descriptions by concatenating the product's name at the beginning of the description for both Unicalss and IFC entities. This step reinforces contextual clarity, as the natural entity names carry significant semantic information. Finally, we manually review and modify the entity descriptions that contain inconsistent information, such as notes related to the schema version history or future depreciation notes.

\subsection{Sampling}
To ensure a robust entity selection when creating task-specific datasets, we implemented the following sampling strategies: For positive sampling, we adopt a semantic diversity approach. Given a targetted subset of built products, we generate text embeddings for all corresponding text inputs, i.e., product names and descriptions. Embeddings are generated using a state-of-the-art text embedding model ("mxbai-embed-large-v1"\citep{li2023angle}). From this set of embeddings, we randomly choose an initial sample as a starting point. Subsequently, we iteratively select additional samples by identifying those that exhibit the slightest similarity to the most recently selected sample, as determined by cosine similarity scores, i.e., the cosine of the angle between two embedding vectors. This process repeats until the desired number of samples is achieved. This method ensures that the samples selected for a particular subset (e.g., products of the same category) yield diverse representations within the embedding space by selecting inputs that are semantically dissimilar to the ones already chosen. For negative sampling, we prioritize the selection of product samples that yield closer semantic similarity to a given query (a product name or description) but belong to a different class. We compute the cosine similarities between the query and negative samples using the same embedding model used in the semantic diversity sampling and select samples with higher similarities. By selecting more similar candidates as negative samples, the dataset can better benchmark the model's capability to capture the subtle differences between closely related classes. This method, commonly known as hard negative sampling, is particularly effective for evaluations involving fine-grained classifications, such as differentiating between closely related categories in IFC and Uniclass classification hierarchies. In all sampling methods, including plain random sampling, once a sample is selected, it is only reused in another subset once all samples included within the pool have been exhausted. This way, we maximize the utilization of available samples and maintain diversity within the datasets.

\section{Benchmark}
\label{sec:bench}
\subsection{Tasks overview}
Evaluating text embeddings across different tasks is crucial for assessing the transferability of their capabilities to various downstream applications. Hence, our proposed benchmark covers three main tasks: clustering, retrieval, and reranking. In addition to domain coverage and cross-task adaptability, evaluating text embedding models requires careful consideration of input text length. To ensure the coverage for varying input lengths, the text entities included in our datasets fall into two categories: (a) sentences, which are derived from product titles/names, and (b) paragraphs, which are derived from product descriptions/definitions. Accordingly, each task-specific dataset in our benchmark is grouped into one of the following categories:

\begin{itemize}
\item \textbf{Sentence to Sentence (S2S):} Utilizing product titles as input text.
\item \textbf{Paragraph to Paragraph (P2P):} Utilizing product descriptions (which can be concatenated with the product name) as input text.
\item \textbf{Sentence to Paragraph (S2P):} Comparing product titles against product descriptions.
\end{itemize}

Our proposed benchmark follows MTEB \citep{muennighoff2022mteb} for reporting text embedding performance scores. Hence, various metrics are implemented within our benchmark, which can be computed with flexible parameter configurations. The primary metrics, which serve as default scores for task-specific as well as overall comparisons reported in this study, are outlined in each task's description.

\subsubsection{Clustering}
Clustering tasks involve grouping similar built products into meaningful clusters based on their similarities in textual representation. Our proposed tasks include S2S and P2P categories, where product names and descriptions act as input text for each dataset type, respectively. Each clustering task dataset is comprised of various subsets, covering diverse subdomain subjects and different levels of granularity. To create the subsets within each clustering dataset, we first select a subset of product labels from one of the three levels of product hierarchy, either from one specific corpus or across both corpora. We then apply the previously described diversity-based sampling method to sample product names (S2S datasets) or descriptions (P2P datasets) for selected labels.

To ensure the quality of the subsets, we evaluate the baseline scores using two embedding models, one for the upper threshold ("mxbai-embed-large-v1"\citep{li2023angle}) and one for the lower threshold ("paraphrase-multilingual-MiniLM-L12-v2"\citep{reimers-2019-sentence-bert}). A subset is included in the dataset only if its score with the upper threshold model is below 0.8  and greater than 1/N with the baseline model, where N is the number of unique labels. The upper and lower thresholds are set to maintain task difficulty and ensure the task performs better than random guessing, respectively. Subsets meeting these criteria are shuffled to eliminate order bias before being added to the dataset.

We compute V-measure scores \citep{rosenberg2007v} by training a mini-batch k-means model using vector embeddings, with k set to the number of unique labels in each clustering subset. The V-measure, ranging from 0 to 1 (higher is better), represents the harmonic mean of two distinct metrics: homogeneity and completeness. Here, homogeneity measures the extent to which clusters contain only products from a single category, while completeness indicates how well all products from a given category are grouped into the same cluster. More details regarding the calculation of V-measure can be found in \citep{rosenberg2007v}. 

\subsubsection{Retrieval} Retrieval tasks aim to identify relevant documents, i.e., product textual descriptions, in response to a given query. Our proposed retrieval datasets are framed as S2P and P2P tasks, where built asset descriptions serve as the corpus (the documents to be retrieved), and product titles and descriptions act as queries for the S2P and P2P tasks, respectively. The query-document relevancy ground truth is derived from existing mappings that identify the alignment between IFC and Unicalss product entities. These mappings, validated and published by NBS\citep{nbs2024}, can be found in the official Unicalss table release\citep{uniclass2024}.

First, we encode all queries and product descriptions into corresponding embedding vectors. These embeddings are then used to calculate the pairwise similarity between a given query and all product descriptions using cosine similarity. Subsequently,  product descriptions included in each retrieval dataset are ranked according to descending cosine similarity scores. Finally, we compute nDCG@10 (Normalized Discounted Cumulative Gain \citep{jarvelin2002cumulated} at rank 10) as the primary metric. This score, which can range between 0 and 1 (higher is better), reflects the relevancy of the ranked products based on their positions within the top 10 ranks by applying a logarithmic discount factor to penalize results that appear lower.

\subsubsection{Reranking} In our reranking tasks, the aim is to rank a set of product descriptions with reference to their relevance to a product query. Similar to retrieval tasks, reranking tasks are framed as S2P and P2P types, and pairwise similarity between query and product description embeddings is computed based on cosine similarity. The primary distinction between retrieval and reranking tasks lies in their scope and focus. While our retrieval tasks involve ranking the entire product corpus, reranking narrows the focus to a smaller set of positive and negative subsets, which are selected using the methods outlined in the previous section to ensure diversity and difficulty (avoiding very high scores from overfitting) within the dataset. Positive and negative samples are selected using the methods described in the previous section, thereby maintaining the diversity and difficulty of the dataset. By concentrating on a smaller and more challenging group of product descriptions, our reranking tasks aim to provide a more fine-grained evaluation of the model's ability to rank relevant items accurately.

Similar to retrieval tasks, we use cosine similarity to compute pairwise similarity between a given query and product descriptions included in corresponding positive and negative sets. Subsequent to ranking the descriptions based on the cosine similarity scores, we compute MAP (Mean Average Precision) as our primary metric. MAP provides an averaged measure of precision across all relevant products, ranging between 0 and 1, with higher values indicating better performance. It is worth noting that retrieval metrics reflect overall ranking quality while reranking metrics focus on how early relevant products appear in the list.

\section{Results}
Table \ref{tab:statistics} provides a comprehensive summary of the dataset statistics across the three main tasks in our benchmark. The unique number of sample entries in our clustering datasets shows that more than half of the samples available from the combined product corpora could pass the quality thresholds explained in the methods section. In the retrieval and reranking task, the same retrieval and reranking document corpus is shared between the subtasks of each task category. This design enables a comparative analysis of model performance on different query types, with S2P focusing on shorter product names and P2P targeting longer product descriptions. We applied a 1:3 positive-to-negative sampling ratio to create a balanced yet challenging evaluation set, ensuring that models must distinguish effectively between relevant and irrelevant documents.

\begin{table*}[t!]
    \centering
    \resizebox{0.7\textwidth}{!}{\begin{tabular}{l|ccccc}
    \toprule
\midrule
\midrule
\multicolumn{1}{l}{} & & & & & \\
\multicolumn{1}{l}{\emph{Clustering tasks}} & No. of & Unique/total & Avg. sample & Total No. of & Avg. unique label \\
\multicolumn{1}{l}{} & subsets & samples & length & unique labels &  per subset \\
\midrule
Clustering-s2s & 18 & 2545/3815 & 28.04 & 31 & 5 \\
Clustering-p2p & 20 & 3067/4577 & 207.91 & 35 & 5 \\
\midrule
\midrule
\multicolumn{1}{l}{} & & & & & \\
\multicolumn{1}{l}{\emph{Retrieval tasks}} & No. of & Avg. query & No. of & Avg. document & No. of document \\
\multicolumn{1}{l}{} &  queries & length & documents &  length &  per query (Avg.) \\
\midrule
Retrieval-s2p & 977 & 30.35 & 2761 & 312.75 & 8 \\
Retrieval-p2p & 977 & 128.5 & 2761 & 312.75 & 8 \\
\midrule
\midrule
\multicolumn{1}{l}{} & & & & & \\
\multicolumn{1}{l}{\emph{Reranking tasks}} & No. of & Avg. query & No. of positives & No. of negatives  & Avg. samples \\
\multicolumn{1}{l}{} & queries & length & (unique/total)  & (unique/total) &  length  \\
\midrule
Reranking-s2p & 179 & 27.89 & 1253/1253 & 2281/3759 & 310.15 \\
Reranking-p2p & 179 & 140.44 & 1253/1253 & 2241/3759 & 309.66 \\
    \bottomrule
    \end{tabular}}
    \caption{Summary of dataset statistics per benchmark task.}
    \label{tab:statistics}
\end{table*}

\begin{figure}[ht]
\centering
\includegraphics[width=0.85\textwidth, trim=0cm 0.5cm 4.5cm 1.5cm, clip]{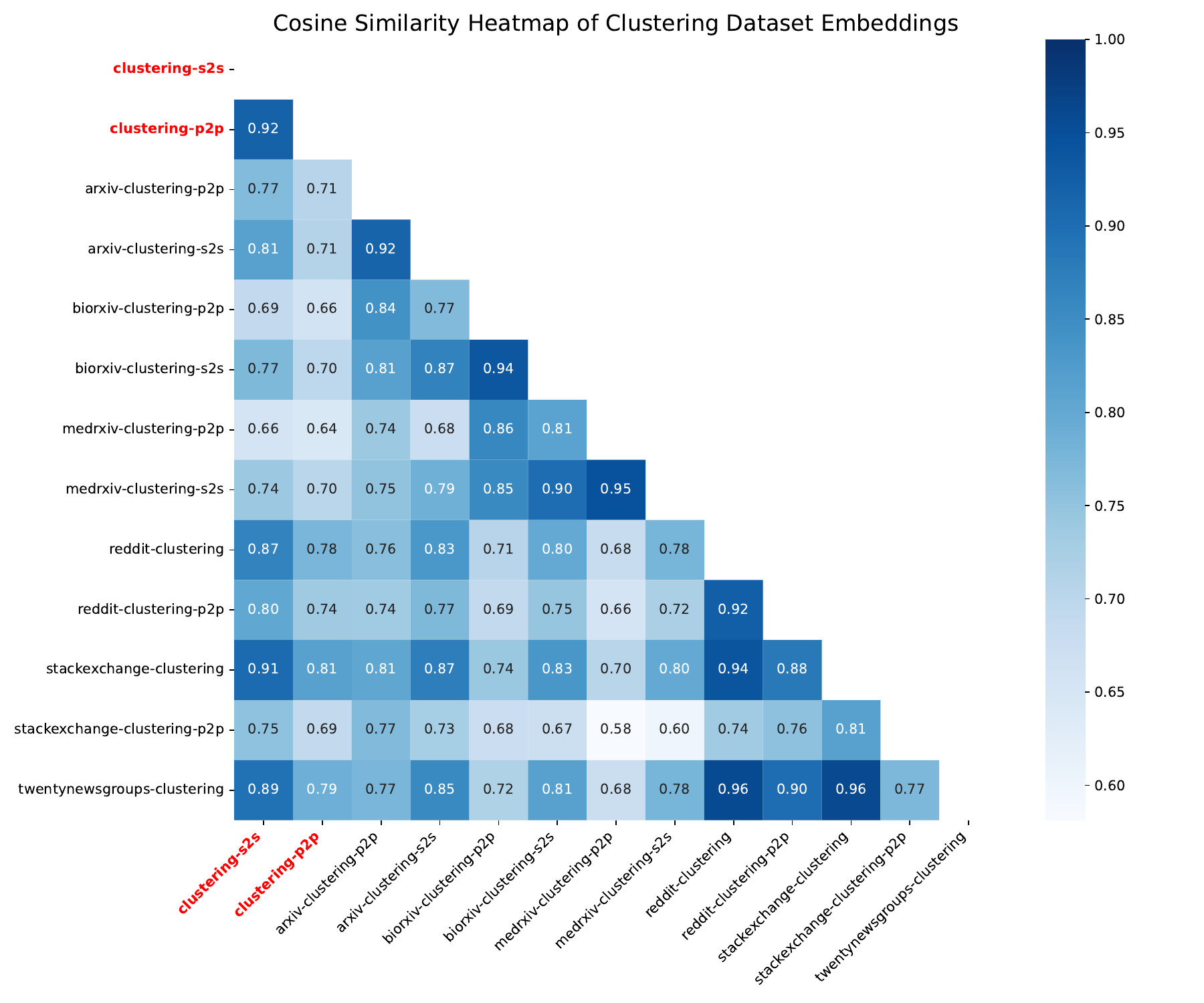}
\caption{Thematic similarity heatmap between our proposed clustering tasks and those from MTEB. Average embeddings are derived from 200 random samples per dataset, encoded using the "mxbai-embed-large-v1" model\citep{li2023angle}. Datasets from our proposed benchmark are highlighted in red.}
\label{fig:data-sim}
\end{figure}

To outline the distinctions between our newly constructed datasets and existing ones, we conducted a thematic semantic similarity comparison between our clustering datasets and those from MTEB benchmark. Using the "stella-en-400M-v5" model, which is the most performant small-sized model in our evaluations (see Table \ref{tab:scores}), we generated embeddings for 200  randomly selected samples and averaged them within each dataset. Figure \ref{fig:data-sim} depicts the cosine similarity matrix as a heatmap, where darker shades indicate higher content similarity. The high similarity scores between our proposed subtasks confirm strong internal consistency within our benchmark. Moreover, moderate to high similarities with StackExchange, Reddit, and Arxiv datasets reflect thematic overlaps with broader domain content. A discussion of the observed similarities is provided in the next section.

\begin{table*}[t!]
    \centering
    \resizebox{\textwidth}{!}{\begin{tabular}{l|ccccccc|cc}
    \toprule
  \multicolumn{1}{r|}{Tasks ($\rightarrow$)} & \multicolumn{2}{c}{Clustering} & \multicolumn{2}{c}{Retrieval} & \multicolumn{2}{c}{Reranking} & Avg. & Param. & MTEB \\
Models ($\downarrow$) & s2s & p2p & s2p & p2p & s2p & p2p & - & (mil) & Rank  \\
\midrule
\midrule
\multicolumn{9}{l}{\emph{Pre-trained without task instructions}}
\\
\midrule
gte-base-en-v1.5&48.38&51.83&79.98&59.42&66.54&66.73&62.15&137&39 \\ 
gte-large-en-v1.5&43.42&51.05&83.32&63.27&\textbf{72.76}&70.15&64.00&434&24 \\ 
bge-base-en-v1.5&43.00&51.78&82.56&61.65&67.01&63.38&61.56&109&43 \\ 
bge-large-en-v1.5&46.69&52.41&82.60&64.86&68.44&65.47&63.41&335&35 \\  
UAE-Large-V1&45.45&49.53&83.32&66.42&70.04&68.53&63.88&335&29 \\ 
GIST-Embedding-v0&46.43&49.96&82.82&62.78&68.81&65.75&62.76&109&41 \\ 
GIST-large-Embedding-v0&47.97&47.91&84.01&67.06&69.53&68.03&64.08&335&34 \\ 
e5-base-v2&42.59&50.24&80.83&61.46&69.11&62.91&61.19&109&64 \\ 
e5-large-v2&42.11&49.45&81.95&64.63&68.61&64.58&61.89&335&55 \\ 
multilingual-e5-large-instruct&48.01&52.82&80.35&64.55&67.85&65.90&63.25&560&42 \\ 
multilingual-e5-small&42.98&48.16&76.38&55.03&64.78&62.34&58.28&118&112 \\ 
all-MiniLM-L12-v2&42.00&46.52&79.97&58.81&66.20&63.97&59.58&33&117 \\ 
paraphrase-multilingual-MiniLM-L12-v2&37.60&45.70&69.01&49.90&61.23&59.15&53.77&118&136 \\ 
gte-base&45.96&51.55&82.91&62.95&68.97&66.26&63.10&109&51 \\ 
gte-large&48.54&55.24&84.32&66.08&70.94&69.25&65.73&335&47 \\ 
gte-small&44.31&55.55&82.37&60.55&68.82&65.23&62.80&33&70 \\ 
\midrule
\multicolumn{9}{l}{\emph{Pre-trained with task instructions}}\\
\midrule
gte-Qwen2-7B-instruct&50.19&\underline{62.39}&\textbf{86.28}&73.20&69.47&67.51&68.17&7069&6 \\ 
mxbai-embed-large-v1&47.49&52.45&83.51&66.60&70.10&69.66&64.97&335&28 \\ 
multilingual-e5-large-instruct&48.10&59.43&82.91&64.42&70.53&69.23&65.77&560&42 \\ 
NV-Embed-v2&\textbf{58.61}&\textbf{67.34}&\underline{85.23}&\textbf{77.02}&66.67&70.34&\textbf{70.87}&7851&1 \\ 
stella-en-1.5B-v5&\underline{53.60}&54.57&84.18&71.21&71.57&\underline{71.77}&67.82&1545&3 \\ 
stella-en-400M-v5&53.39&55.78&84.60&70.00&69.58&69.36&67.12&435&7 \\
\midrule
\multicolumn{9}{l}{\emph{Proprietary embedding APIs}}\\
\midrule
text-embedding-3-small & 49.72 & 49.72  & 79.97	& 65.68 & 65.33 &	66.99 & 62.90 & - & 58 \\
text-embedding-3-large & 49.75 & 55.48  & 84.99 & \underline{75.38} & \underline{71.93} & \textbf{72.46} & \underline{68.33} & - & 30\\
    \bottomrule
    \end{tabular}}
    \caption{Average scores of benchmarked models per task, based on the task-specific metrics mentioned in the task descriptions. The first and second highest scores for each task are highlighted in bold and underlined, respectively. MTEB ranks are sourced from records as of September 21, 2024.}
    \label{tab:scores}
\end{table*}

In our benchmarking experiments, we evaluated models across a broad range of sizes, from relatively small models with 33 million parameters to significantly larger models exceeding seven billion parameters. However, due to computational constraints, the majority of models tested have less than one billion parameters. The selected models span various positions on the most recent record of MTEB leaderboard (as of September 21, 2024), ranging from first place (i.e., "NV-Embed-v2"\citep{lee2024nv}) to 136th place (i.e., "paraphrase-multilingual-MiniLM-L12-v2"). For models that are pre-trained with instruction-based data, we used built-in or recommended prompts as provided in the model card's official web page or associated research papers, when available. For example, "mxbai-embed-large-v1" requires custom prompts only for retrieval and reranking tasks, while "NV-Embed-v2" needs specific task-based prompts for clustering tasks as well. For models without built-in task instructions, we applied a general set of prompts to ensure consistency across tasks (prompts are available at the project's public GitHub repository\citep{benchGithub}). 

The top-ranked model in our benchmark, "NV-Embed-v2", also holds first place on the latest MTEB leaderboard. However, it does not consistently outperform all other models across all tasks. In fact, a closer examination reveals variability in model size and performance relationship. For example, "gte-small", the smallest model in our evaluation with 33 million parameters, delivers competitive scores, nearly matching the average scores of models ten times its size and even outperforming larger models in specific tasks. Despite the previously reported strong correlation between model size and performance\citep{muennighoff2022mteb}, our experiments show that superior performance associated with larger models is only evident at the extreme upper end of the parameter scale. This observation supports the growing emphasis on developing and deploying smaller, more efficient models for both research and real-world applications in this specialized field.

Motivated by the hypothesis that existing datasets with similar thematic content would yield comparable performance evaluations, we examined the consistency of relative model performances as follows: Given the observed thematic similarity between our clustering datasets and specific MTEB datasets, particularly "StackExchange" and "Reddit" (see Figure \ref{fig:data-sim}), we compared the rankings of model performance across both our datasets and the selected MTEB datasets. As it can be seen from Table \ref{tab:clus-ranks}, the comparative evaluation of the relative rankings indicates a notable variation in model performances, notably in the case of "multilingual-e5-large-instruct", "gte-small", "stella\_en\_1.5B\_v5", and "text-embedding-3-small". These observed variabilities further highlight the limitations of relying on general-purpose benchmark datasets, even when relatively high thematic similarities are present, underscoring the importance of domain-specific evaluations.

While our benchmarking experiments primarily focused on open-source models, we also included the proprietary text embedding models from OpenAI, both the small and large versions. The inclusion of the proprietary models is motivated by a recent study where closed-source models tend to achieve relatively higher performance when embedding text in underrepresented languages \citep{enevoldsen2024scandinavian}. We hypothesize that built asset text, as an underexplored domain, might be similarly better represented by proprietary models. Notably, text-embedding-3-large ranks second in our benchmark, performing nearly on par with the top-ranked model. In contrast, the smaller model performed more moderately, ranking in the middle of our benchmark. While the former observation aligns with the findings of \citep{enevoldsen2024scandinavian}, the latter is in line with the latest MTEB leaderboard results where closed-source commercial
embedding APIs generally underperform compared to their open-source counterparts. These observations raise questions about the underlying factors. However, the lack of knowledge about the key characteristics of proprietary models, such as their size and diversity in training data, prevents us from offering a detailed, conclusive account of their relative performance. 

Our benchmarking results reveal a notable difference in performance between shorter and longer text inputs in different tasks. In particular, across the board, models consistently show lower performance in the S2S clustering task compared to the P2P one. This observation can be attributed to the limited presence of contextual clues given the significantly short length of the input text in the S2S clustering task (see Table \ref{tab:statistics}). On the other hand, in reranking and retrieval tasks, the majority of the models yield moderately higher scores in S2P tasks. The likely explanation for the latter observation is that the shorter length of the sentences (product names) in S2P tasks can lead to a lower amount of irrelevant information (noise) in the input query. Since product names tend only to encapsulate the critical information about the target product, they can yield more precise and discriminative text (query) representations for similarity matching.

\begin{table}[htbp]
\centering
\begin{adjustbox}{max width=\textwidth}
\small
\begin{tabular}{lccccccccccccccccccccccc}
\toprule
 & \rotatebox{90}{NV-Embed-v2} & \rotatebox{90}{gte-Qwen2-7B-instruct} & \rotatebox{90}{multilingual-e5-large-instruct} & \rotatebox{90}{stella\_en\_400M\_v5} & \rotatebox{90}{gte-small} & \rotatebox{90}{text-embedding-3-large} & \rotatebox{90}{gte-large} & \rotatebox{90}{stella\_en\_1.5B\_v5} & \rotatebox{90}{mxbai-embed-large-v1} & \rotatebox{90}{bge-large-en-v1.5} & \rotatebox{90}{gte-base-en-v1.5} & \rotatebox{90}{bge-base-en-v1.5} & \rotatebox{90}{gte-base} & \rotatebox{90}{gte-large-en-v1.5} & \rotatebox{90}{e5-base-v2} & \rotatebox{90}{GIST-Embedding-v0} & \rotatebox{90}{text-embedding-3-small} & \rotatebox{90}{UAE-Large-V1} & \rotatebox{90}{e5-large-v2} & \rotatebox{90}{multilingual-e5-small} & \rotatebox{90}{GIST-large-Embedding-v0} & \rotatebox{90}{all-MiniLM-L12-v2} & \rotatebox{90}{paraphrase-multilingual-MiniLM-L12-v2} \\
\midrule
clustering-p2p (ours) & 1 & 2 & 3 & 4 & 5 & 6 & 7 & 8 & 9 & 10 & 11 & 12 & 13 & 14 & 15 & 16 & 17 & 18 & 19 & 20 & 21 & 22 & 23 \\
stackexchange-clustering & 1 & 3 & 10 & 4 & 19 & 7 & 9 & 2 & 16 & 14 & 8 & 20 & 11 & 6 & 18 & 15 & 5 & 12 & 17 & 21 & 13 & 22 & 23 \\
reddit-clustering & 4 & 1 & 16 & 3 & 18 & 10 & 6 & 2 & 9 & 15 & 14 & 21 & 12 & 11 & 17 & 13 & 5 & 8 & 19 & 23 & 7 & 20 & 22 \\
\bottomrule
\end{tabular}
\end{adjustbox}
\caption{Comparison of model rankings across datasets with high thematic similarity (see Figure \ref{fig:data-sim})}
\label{tab:clus-ranks}
\end{table}

\section{Discussion}
Our benchmarking results offer critical insights into the effectiveness of state-of-the-art pre-trained text embedding models in aligning built asset information. One of the key findings of our study is the variability in performance across tasks, even among top-performing models. Our results suggest that model effectiveness is not strongly correlated across model sizes, emphasizing that size alone is not a reliable predictor of model performance in the specialized domain of built asset information management. The interpretation of the relationship between model size and embedding effectiveness is further complicated by the performance gap observed when comparing models pre-trained with and without instruction tuning. Instruction-tuned models showed higher performance in the majority of our benchmark tasks. Considering the larger size of the instruction-tuned models included in our experiments, the latter observation raises an important question for future research: To what extent can instruction-tuning help smaller models adapt to the specialized domain of the built environment? This opens a promising line of investigation into how task-specific training with instruction-based data can better align a model’s understanding with the intricate semantics of built asset data, particularly for models with smaller sizes. Finally, in addition to the variability in model performance across different tasks and text input lengths, the results of our comparative examinations highlight the limited transferability of evaluations based on general benchmarks. Our experiments indicate that, even with relatively high thematic similarity, general-purpose benchmarks remain inadequate in capturing the unique semantic complexity and contextual dependencies present in the textual descriptions of the built asset.

The above-mentioned points highlight the critical need for tailored benchmarking datasets to examine the effectiveness of various domain adaptation strategies in this field of research. Our work contributes to the body of research by laying a robust foundation for future evaluations and providing a benchmark that is carefully constructed to reflect the complexities of built asset data. Our proposed datasets cover diverse subdomains and exhibit varying levels of granularity, mirroring real-world scenarios where built products are required to be mapped across various data dictionaries. The datasets can be used not only for evaluating new or fine-tuned text embedding models for cross-mapping built asset data but also as a contextually rich text corpus to support the training of task-specific language models for other downstream tasks, such as information extraction. Finally, this work contributes to the broader discourse on the transferability of the general-purpose language models' capabilities by focusing on built asset data as a representative example of niche and underexplored domains.

One key limitation of our study is that the text sources used in our work are exclusively in English, limiting the generalizability of our findings to other languages. Another significant challenge was identifying data sources that were both of high quality and could be redistributed as public datasets. In this light, although the developed datasets proved sufficient for our current analysis, future work could benefit from larger-scale datasets and introduce training and validation splits to support new tasks. It is recommended to prioritize exploring more extensive and diverse text sources to include multiple languages and new tasks where the availability of large training splits plays a crucial role, such as text classification or reranking based on cross-encoder architectures. Finally, through the public release of our benchmarking resources in alignment with the MTEB's open-source software, we aim to ensure the reproducibility and extendability of our work through community-driven enhancements. 

\section*{Data availability}
The datasets and codes developed in this study are openly accessible at the following GitHub repository: \url{https://github.com/mehrzadshm/built-bench-paper}. All materials are licensed under the Creative Commons Attribution-NoDerivatives 4.0  International License (CC BY-ND 4.0). Any future updates, including references to additional data and relevant resources, will be incorporated into this repository.

{
\small
\bibliography{references}
}

\end{document}